\documentclass[twoside,11pt]{article}
\usepackage[preprint]{jmlr2e}
\usepackage{blindtext}
\usepackage{amsfonts}
\usepackage{amsmath}
\usepackage{listings}
\usepackage[table]{xcolor}
\usepackage{hyperref}
\hypersetup{
  colorlinks=true,
  linkcolor=dark-blue,
  urlcolor=dark-blue,
  citecolor=dark-blue,
}

\usepackage{listings}
\usepackage{xcolor}

\definecolor{dark-blue}{rgb}{0.15,0.15,0.4}
\definecolor{codegreen}{rgb}{0,0.6,0}
\definecolor{codegray}{rgb}{0.5,0.5,0.5}
\definecolor{codepurple}{rgb}{0.58,0,0.82}
\definecolor{backcolour}{rgb}{0.95,0.95,0.92}

\lstdefinestyle{mystyle}{
    backgroundcolor=\color{backcolour},   
    commentstyle=\color{codegreen},
    keywordstyle=\color{magenta},
    numberstyle=\tiny\color{codegray},
    stringstyle=\color{codepurple},
    basicstyle=\ttfamily\scriptsize\color{blue!30!black},    emph={int,char,double,float,unsigned,void,bool},
    emphstyle={\color{blue}},
    morekeywords={>,<,.,;,-,!,=,~},
    otherkeywords={>,<,.,;,-,!,=,~},
    breakatwhitespace=false,         
    breaklines=false,             
    language=python,
    captionpos=b,                    
    keepspaces=true,                   
    showspaces=false,                
    showstringspaces=false,
    showtabs=false,                  
    tabsize=4
}

\definecolor{bgcolor}{HTML}{E0E0E0}
\let\oldtexttt\texttt

\renewcommand{\texttt}[1]{
  \colorbox{bgcolor}{\oldtexttt{#1}}
  }

\newcommand{\Fortuna}{\textsf{Fortuna}}

\usepackage{xcolor}

\usepackage{lastpage}
\jmlrheading{23}{2022}{1-\pageref{LastPage}}{x/xx; Revised x/xx}{x/xx}{21-0000}{Gianluca Detommaso, Alberto Gasparin, Michele Donini, Matthias Seeger, Andrew Gordon Wilson and Cedric Archambeau}

\ShortHeadings{\Fortuna: A Library for Uncertainty Quantification in Deep Learning}{Detommaso, Gasparin, Donini, Seeger, Wilson and Archambeau}
\firstpageno{1}

\begin{document}

\title{Fortuna: A Library for Uncertainty Quantification in Deep Learning}
\author{\name Gianluca Detommaso$^{1}$ \email detomma@amazon.de
       \AND
       \name Alberto Gasparin$^{2}$ \email albgas@amazon.de
       \AND
       \name Michele Donini$^{1}$ \email donini@amazon.de
       \AND
       \name Matthias Seeger$^{1}$ \email matthis@amazon.de
       \AND
       Andrew Gordon Wilson$^{3}$ \email wilsmman@amazon.com
       \AND
       \name Cedric Archambeau$^{1}$ \email cedrica@amazon.de
       \AND
       \addr ~$^{1}$AWS, Berlin, Germany\\
       \addr ~$^{2}$Amazon, Berlin, Germany\\
       \addr ~$^{3}$AWS\ \&\ New\ York\ University
       }

\editor{} %\agw{no editor assigned yet; submitted, revised, etc., also needs to be updated to `xx'}}
% \date{November 2022}

\maketitle

\begin{abstract}
    We present \Fortuna, an open-source library for uncertainty quantification in deep learning. Fortuna supports a range of calibration techniques, such as conformal prediction that can be applied to any trained neural network to generate reliable uncertainty estimates, and scalable Bayesian inference methods that can be applied to Flax-based deep neural networks trained from scratch for improved uncertainty quantification and accuracy. By providing a coherent framework for advanced uncertainty quantification methods, \Fortuna~simplifies the process of benchmarking and helps practitioners build robust AI systems.
    % AGW notes:
    % - More concrete detail 
    % - Vision of the library (e.g., sota methods, coherent framework, while retaining focus)
\end{abstract}

\section{Introduction}\label{sec:intro}
%Accurate estimation of predictive uncertainty is essential in applications that involve important decision-making. 
Virtually every application of machine learning ultimately involves decision making under uncertainty. %In order to make more robust decisions, we need estimates of the predictive uncertainty.
Predictive uncertainty %can be used to 
lets us evaluate the trustworthiness of model predictions, prompts human intervention, or determines whether a model can be safely deployed in the real-world. Proper uncertainty estimation is crucial for ensuring the reliability and safety of machine learning applications.

Unfortunately, deep neural networks are often overconfident. In classification, overconfidence means that the estimated probability of the predicted class is significantly higher than the actual proportion of correctly classified input data points \citep{guo2017calibration}. Overconfidence is problematic because impact decisions and ensuing actions. For example, a doctor may have requested an additional test if she were to know that a diagnosis made an AI was less confident and a self-driving car may have asked a human driver to take over if it was unsure about the existence of an obstacle in front of the car. 
Hence, calibrated uncertainty estimates are vital for assessing the reliability of %model predictions
machine learning systems, triggering human intervention, or judging whether a model can be safely deployed.

There are %numerous published techniques
many techniques for estimating %the predictive uncertainty %of a prediction or
and %for 
calibrating %such 
uncertainty estimates, including temperature scaling \citep{guo2017calibration}, conformal prediction %methods
\citep{vovk2005algorithmic} and Bayesian inference \citep{wilson2020bayesian}. While there are existing open-source implementations of methods for uncertainty quantification \citep{nado2021uncertainty, chung2021uncertainty, uq360-june-2021, phan2019composable, bingham2019pyro}, they tend to provide general-purpose probabilistic programming languages, without support for scalable state-of-the-art methods, or individual implementations that do not support a broad range of methods in a unified interface.
While modern techniques are highly scalable and practical, the lack of a unified framework has hindered the adoption of uncertainty quantification in practice, 
 a gap that we aim to address by the release of \Fortuna.

\Fortuna~is an open-source library for uncertainty quantification that brings together state-of-the-art scalable methods from the literature and provides them to users through a standardized and easy-to-use interface.\footnote{\Fortuna~\href{https://aws-fortuna.readthedocs.io/en/latest/}{documentation} and \href{https://github.com/awslabs/fortuna}{GitHub repository}.} Our goal is to make it simpler for practitioners to deploy a variety of advanced uncertainty quantification techniques. \Fortuna~allows users to calibrate uncertainty estimates for trained deep neural networks using techniques such as temperature scaling and conformal prediction. \Fortuna~ 
also provides scalable Bayesian inference methods for the training of neural networks from scratch, which can improve both calibration and accuracy. 
\Fortuna~is written in \href{https://jax.readthedocs.io/en/latest/}{\textsf{JAX}} \citep{jax2018github}, a fast growing NumPy-like framework that 
allows for native and efficient computation of gradients essential for %many 
large-scale Bayesian inference,
and adopts \href{https://flax.readthedocs.io/en/latest/}{\textsf{Flax}} \citep{flax2020github}, which has been integrated in many other AI frameworks, including \href{https://huggingface.co/docs}{\textsf{Hugging Face}} \citep{wolf-etal-2020-transformers}. 
In Appendix~\ref{app:experiments}, we %present a comparison of 
compare the calibration of uncertainty estimates for models trained via a standard regularized MLE estimation procedure versus \Fortuna's default setting on several computer vision %datasets
tasks. 
\Fortuna~improves the accuracy of uncertainty estimates across various metrics, including the predictive log likelihood, the expected calibration error (ECE) \citep{guo2017calibration}, and the Brier score \citep{brier1950verification}, on multiple datasets such as Cifar10, Fashion MNIST, MNIST, MNIST corrupted, and SVHN. Thus using \Fortuna~can easily help practitioners calibrate their uncertainty estimates.

\section{Usage modes}
\begin{figure}
    \centering
    \includegraphics[scale=0.45, trim={0cm 0.1cm 0cm 0.1cm}, clip]{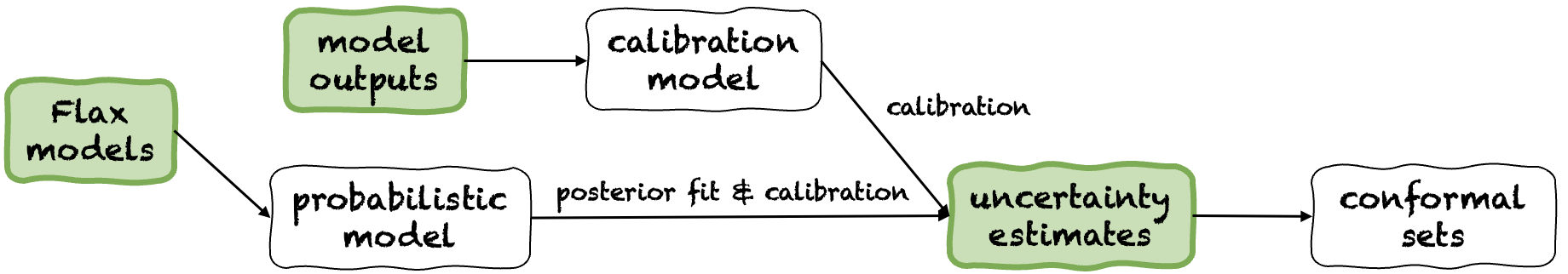}
    \caption{\Fortuna~provides three usage modes, each starting from one of the colored panels.}
    \label{fig:my_label}
    \vspace{-0.92cm}
\end{figure}

One can make use of \Fortuna~offers in three ways: one can starting (1) from uncertainty estimates (Section \ref{sec:uncertainty}), (2) from model outputs (Section \ref{sec:outputs}), or (3) from \textsf{Flax} models (Section \ref{sec:flax}). These modes are ordered in terms of decreasing convenience, but increasing flexibility and performance. We illustrate these usage modes in Figure~\ref{fig:my_label}.

\subsection{Starting from uncertainty estimates}\label{sec:uncertainty}
Starting from uncertainty estimates is the easiest method of interacting with the library. This usage mode offers conformal prediction methods for both classification and regression tasks. These methods take uncertainty estimates in the form of a \texttt{numpy.ndarray} and return a set of predictions with a user-specified probability level. In univariate regression tasks, conformal sets can be thought of as confidence or credible intervals with a calibration guarantee. However, if the provided uncertainty estimates are inaccurate, the resulting conformal sets can be too large to be useful. For this reason, we recommend the usage modes in \ref{sec:outputs} and \ref{sec:flax} to obtain calibrated uncertainty estimates.

At the time of writing, for classification the library supports  %both 
the baseline conformal prediction by \cite{vovk2005algorithmic} and the more advanced adaptive conformal prediction by \cite{romano2020classification}. %, which can work better. 
For regression, it offers conformalized quantile regression \citep{romano2019conformalized}, conformal intervals from a scalar uncertainty measure \citep{angelopoulos2022image}, Jackknife+, Jackknife-minmax and CV+ \citep{barber2021predictive}. The method of choice depends may depend on which uncertainty estimates the user wants to start from, or whether cross-validation methods are feasible in their setting.

\paragraph{Example.} We wish to calibrate credible intervals with coverage error given by \texttt{error}. We assume to be given credible intervals (\texttt{test\_lower\_bounds} and \texttt{test\_upper\_bounds}) corresponding to different test input variables, and credible intervals for several validation inputs (\texttt{val\_lower\_bounds} and \texttt{val\_upper\_bounds}), along with corresponding validation targets (\texttt{val\_targets}). The following code produces conformal prediction intervals as calibrated versions of the test credible intervals.

\begin{lstlisting}[language=python]
from fortuna.conformal.regression import QuantileConformalRegressor
conformal_intervals = QuantileConformalRegressor().conformal_interval(
     val_lower_bounds=val_lower_bounds, val_upper_bounds=val_upper_bounds,
     test_lower_bounds=test_lower_bounds, test_upper_bounds=test_upper_bounds,
     val_targets=val_targets, error=error)
\end{lstlisting}

\subsection{Starting from model outputs}\label{sec:outputs}
This mode assumes that a %deterministic 
model has already been trained, possibly in another framework, and that model outputs for each input data point are available to %use in
\Fortuna~(i.e., estimates of logits for classification or conditional expectations and variances for regression). This usage mode allows for calibration of the model outputs, estimation of the uncertainty, computation of the metrics, and generation of the conformal sets. Compared to the mode in Section \ref{sec:uncertainty}, this mode offers additional control over the final uncertainty estimates, at the price of a few lines more code. Nevertheless, if the model was trained as a point estimator, the resulting epistemic uncertainty estimates may not be reliable, and the usage mode in Section \ref{sec:flax} may be even more preferable.

Models outputs are passed to an additional model written in \textsf{Flax}. The simplest and most popular choice is temperature scaling, which scales the outputs of the model (specifically, the logits in classification and the output uncertainty in regression) %variances in regression) 
using a single parameter \citep{guo2017calibration}. The temperature scaling method is provided explicitly in \Fortuna.

\paragraph{Example.} Assuming we have calibration and test model outputs (\texttt{calib\_outputs} and \texttt{test\_outputs}) as well as calibration targets (\texttt{calib\_targets}), the following code provides a minimal example of obtaining calibrated predictive entropy estimates for a classification task.

\begin{lstlisting}[language=python]
from fortuna.calib_model import CalibClassifier
calib_model = CalibClassifier()
status = calib_model.calibrate(calib_outputs=calib_outputs, 
                               calib_targets=calib_targets)
test_entropies = calib_model.predictive.entropy(outputs=test_outputs)
\end{lstlisting}

\subsection{Starting from \textsf{Flax} models}\label{sec:flax}
The usage modes discussed above are agnostic to how model outputs or initial uncertainty estimates are obtained. By contrast, the most powerful mode discussed here requires the input of deep learning models written in \href{https://flax.readthedocs.io/en/latest/index.html}{\textsf{Flax}}. By replacing traditional model training with scalable Bayesian inference, one can improve the quantification of predictive uncertainty and accuracy significantly \citep{wilson2020bayesian}. Bayesian methods can represent \textit{epistemic} uncertainty --- uncertainty in the model parameters due to lack of information. Since neural networks can represent many different compelling solutions through different parameter settings, Bayesian methods can be particularly useful in deep learning. \Fortuna~offers various scalable Bayesian inference methods that provide uncertainty estimates, as well as improved accuracy and calibration, at the price of a training time overhead.

At the time of writing, \Fortuna~supports  %the following Bayesian inference methods:
Maximum-A-Posteriori (MAP) \citep{bassett2019maximum}, Automatic Differentiation Variational Inference (ADVI) \citep{kucukelbir2017automatic}, Deep Ensembles \citep{wilson2020bayesian}, Laplace approximation with diagonal Generalized Gauss-Newton (GNN) Hessian approximation \citep{daxberger2021laplace, schraudolph2002fast} and SWAG \citep{maddox2019simple}.

\paragraph{Example.} 
If we have a \textsf{Flax} deep learning classifier %classification deep learning model 
(\texttt{model}) that maps inputs to \texttt{output\_dim} logits, as well as training, validation, and calibration TensorFlow data loaders (\texttt{train\_data\_loader}, \texttt{val\_data\_loader}, \texttt{test\_data\_loader}), the following code provides a minimal example for obtaining calibrated probability estimates. %for a classification task.

\begin{lstlisting}[language=python]
from fortuna.data import DataLoader
train_data_loader = DataLoader.from_tensorflow_data_loader(train_data_loader)
calib_data_loader = DataLoader.from_tensorflow_data_loader(val_data_loader)
test_data_loader = DataLoader.from_tensorflow_data_loader(test_data_loader)
test_inputs_loader = test_data_loader.to_inputs_loader()

from fortuna.prob_model import ProbClassifier
prob_model = ProbClassifier(model=model)
status = prob_model.train(train_data_loader=train_data_loader, 
                          calib_data_loader=calib_data_loader)
test_means = prob_model.predictive.mean(inputs_loader=test_inputs_loader)
\end{lstlisting}

\section{Conclusion}
We introduced \Fortuna, a library for uncertainty quantification in deep learning. \Fortuna~supports %brings together 
state-of-the-art methods in a 
coherent interface. To get started with \Fortuna, you can consult the \href{https://github.com/awslabs/fortuna}{GitHub repository}, the \href{https://aws-fortuna.readthedocs.io/en/latest/}{documentation}, and the \href{https://aws-preview.aka.amazon.com/blogs/machine-learning/introducing-fortuna-a-library-for-uncertainty-quantification/}{AWS blog post}.

\bibliography{biblio}

\appendix

\section{Classification using \Fortuna}\label{app:experiments}
In this section, we compare the calibration of we compare standard training procedure (regularized cross-entropy loss minimization) and the default options in \Fortuna~(SWAG \citep{maddox2019simple} plus temperature scaling \citep{guo2017calibration}) over several calibration data sets and computer vision classification tasks. In order to train MNIST \citep{lecun1998gradient}, we use a two-layer MLP model \citep{haykin2004comprehensive}; for SVHN \citep{Netzer2011} and MNIST corrupted we took a LeNet5 \citep{lecun1998gradient}; for Fashion MNIST \citep{xiao2017fashion} we chose a ResNet50 \citep{he2016deep}; finally, for Cifar10 \citep{Krizhevsky09learningmultiple}, we take a WideResNet28-18 \citep{zagoruyko2016wide}. In all experiments, for the standard training procedure we took an Adam optimizer \citep{kingma2014adam} with learning rate of 1e-3, running for 300 epochs. SWAG was run post-hoc of standard training with the same Adam optimizer for 300 epochs. Temperature scaling used an Adam optimizer with learning rate of 1e-1, also for 300 epochs. 

\begin{figure}[h!]
    \centering
    \includegraphics[width=\textwidth, trim={0.1cm, 0.1cm, 0.1cm, 0.1cm}, clip]{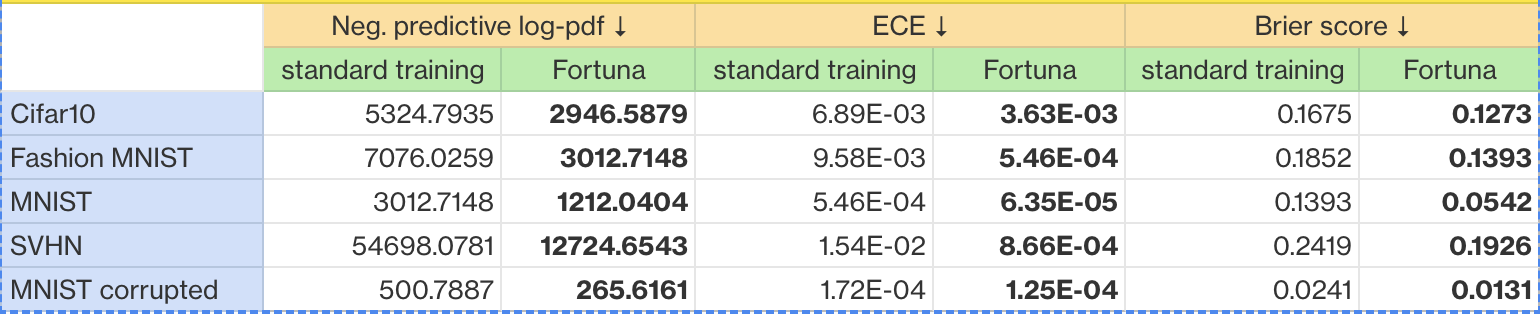}
    \caption{A comparison of calibration metrics obtained with a standard training procedure and with \Fortuna's default options shows that \Fortuna~consistently improves calibration across all metrics and datasets.}
    \label{fig:experiments}
\end{figure}
The results in Table \ref{fig:experiments} show that using \Fortuna~out-of-the-box leads to significantly better calibration compared to standard training methods. All metrics consistently improved across all data sets when using \Fortuna. While further optimization such as fine-tuning hyperparameters, choosing more advanced architectures, and using more suitable optimizers may yield even better results, our focus was on demonstrating how \Fortuna~can easily and reliably improve the accuracy of uncertainty estimates.

\end{document}